\documentclass[acmtog]{acmart}
\acmSubmissionID{xxxx}

\usepackage{booktabs} 

\usepackage{url}
\usepackage{xcolor}

\usepackage[ruled]{algorithm2e} 

\SetAlFnt{\small}
\SetAlCapFnt{\small}
\SetAlCapNameFnt{\small}
\SetAlCapHSkip{0pt}

\acmJournal{TOG}

\title{DiffBrush \raisebox{-1mm}{\includegraphics[scale=0.08]{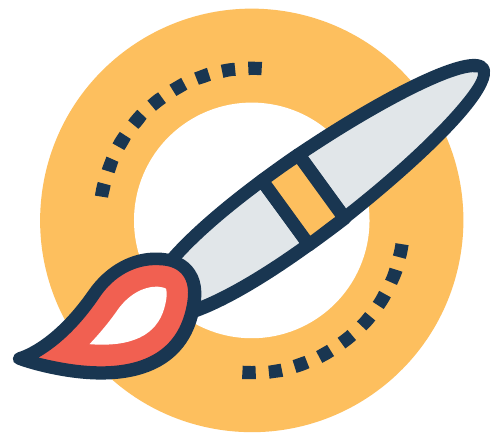}} : Just Painting the Art by Your Hands}

\begin{document}

\author{Jiaming Chu}
\orcid{0000-0002-3040-3674}
\affiliation{
  \institution{Beijing University of Post and Telecommunication}
  \city{Beijing}
  \country{China}}
\email{chuiaming886@bupt.edu.cn}

\author{Lei Jin}
\orcid{0000-0003-4855-2464}
\affiliation{
\institution{Beijing University of Posts and Telecommunications}
\city{Beijing}
\country{China}
}
\email{jinlei@bupt.edu.cn}

\author{Tao Wang}
\affiliation{
\institution{Beijing University of Posts and Telecommunications}
\city{Beijing}
\country{China}
}
\email{wangtao@bupt.edu.cn}

\author{Junliang Xing}
\affiliation{
\institution{Tsinghua University}
\city{Beijing}
\country{China}
}
\email{jlxing@tsinghua.edu.cn}

\author{Jian Zhao}
\affiliation{
\institution{Institute of AI (TeleAI), China Telecom and the School of Artificial Intelligence}
\department{EVOL Lab}
\city{Beijing}
\country{China}
}
\affiliation{
\institution{Northwestern Polytechnical University (NWPU)}
\department{Optics and Electronics (iOPEN)}
\city{Beijing}
\country{China}
}
\email{zhaoj90@chinatelecom.cn}


\begin{abstract}
    The rapid development of image generation and editing algorithms in recent years has enabled ordinary user to produce realistic images. However, the current AI painting ecosystem predominantly relies on text-driven diffusion models (T2I), which pose challenges in accurately capturing user requirements. Furthermore, achieving compatibility with other modalities incurs substantial training costs. To this end, we introduce DiffBrush, which is compatible with T2I models and allows users to draw and edit images. By manipulating and adapting the internal representation of the diffusion model, DiffBrush guides the model-generated images to converge towards the user's hand-drawn sketches for user's specific needs without additional training. DiffBrush achieves control over the color, semantic, and instance of objects in images by continuously guiding the latent and instance-level attention map during the denoising process of the diffusion model. 
    Besides, we propose a latent regeneration, which refines the randomly sampled noise in the diffusion model, obtaining a better image generation layout.
    Finally, users only need to roughly draw the mask of the instance (acceptable colors) on the canvas, DiffBrush can naturally generate the corresponding instance at the corresponding location. 
\end{abstract}

%
%


%
%


\begin{teaserfigure}
  \includegraphics[width=\textwidth]{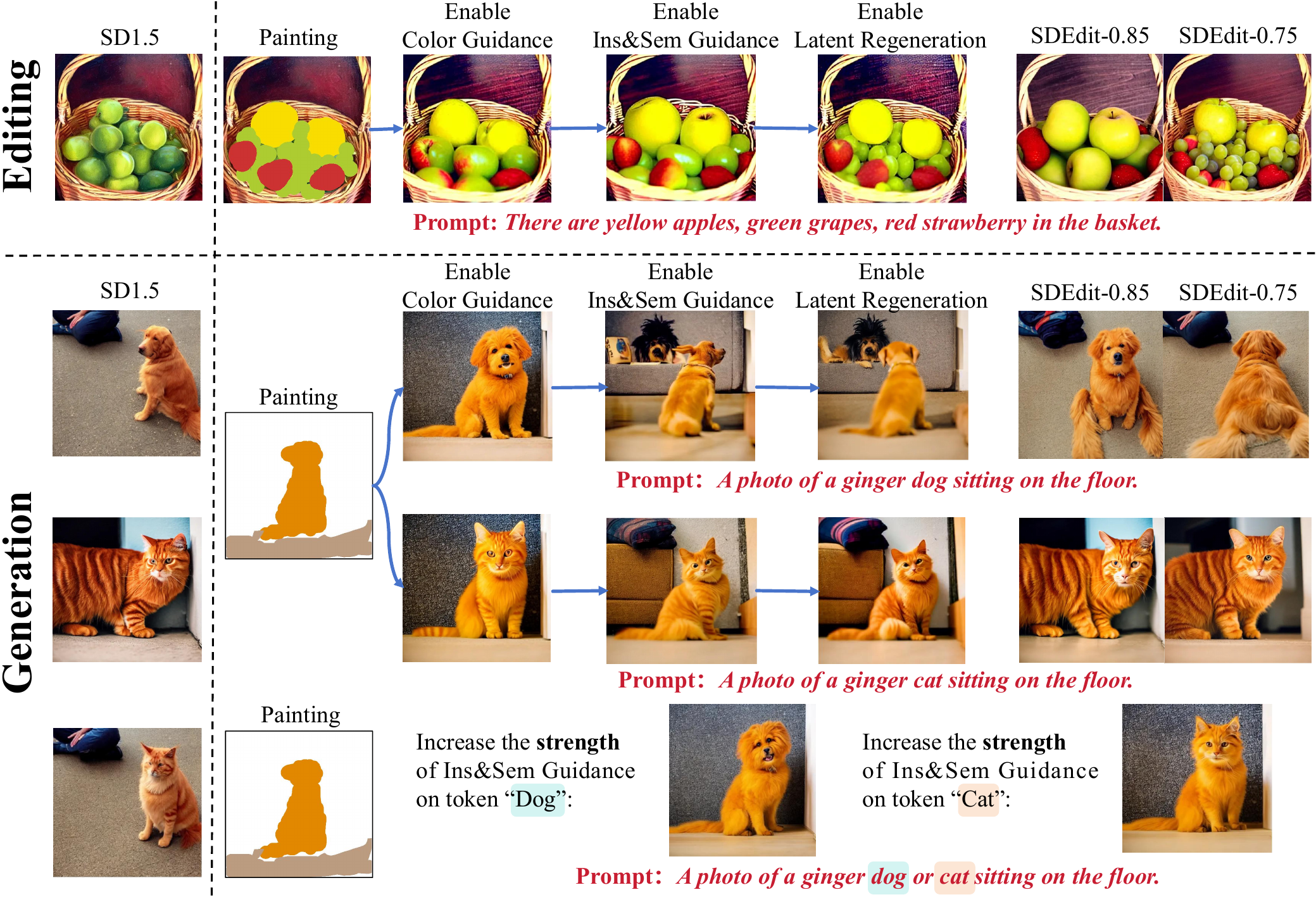}
  \vspace{-0.4cm}
  \caption{The motivation of DiffBrush. DiffBrush, which is a training-free method, provides color, instance, and semantic control, and can refine the initial noise distribution through rough sketches drawn by users.}
  \label{fig:first_fig}
\end{teaserfigure}
\maketitle

\section{Introduction}

The rapid development of image generation~\cite{songdenoising, songscore, ho2020denoising, ramesh2022hierarchical, rombach2022high, podellsdxl} in recent years has had a great impact, narrowing the gap between ordinary people and professional artists, and allowing ordinary people to produce vivid painting. Among them, the text-to-image (T2I) models~\cite{ramesh2022hierarchical, rombach2022high, podellsdxl, flux} has become the mainstream of AI painting, and there are many innovations and works based on T2I that are changing the ecology of AI painting~\cite{civitai, huggingface}. However, finding a text prompt that accurately describes user needs is still a very difficult task. For T2I models, the diffusion model maps standard normal distribution Gaussian noise to real images with a one-to-one relationship under a fixed prompt. Nevertheless, the sampled real images may not fully meet the user's requirements, and user can only approach the his target image by continuously modifying prompts or irregularly filtering random seeds, which may consume a lot of time and energy. 

To address these issues, many researchers have proposed conditional image generation~\cite{dhariwal2021diffusion, ho2022classifier, bansal2023universal, graikos2022diffusion,zhang2023adding, yu2022scaling, meng2021sdedit} and editing algorithms~\cite{brooks2023instructpix2pix, chen2024training, gal2022image, hertzprompt, kingma2021variational, kwondiffusion, liu2022compositional}. However, the majority of these methods rely on diverse generation models for fine-tuning and aligning the feature space, necessitating users to bear substantial training costs, such as ControlNet~\cite{zhang2023adding}. A small number of training-free methods cannot simultaneously balance color and instance accuracy, such as SDEdit~\cite{meng2021sdedit}, which can generate color matched images using a T2I model by hand drawing blurry sketches by users. But for adjacent objects with similar colors, they can only rely on the precise description of text prompt, which is particularly prone to target confusion. Image-editing algorithms such as Self-Guidance~\cite{epstein2023diffusion} or MaskCtrl~\cite{cao2023masactrl} require a reference image as a basis to generate new images. Although specific instances can be edited through conditions such as mask, editing the appearance and other attributes of the objects heavily relies on text descriptions or additional reference images, which causes inconvenience for ordinary users using image generation.

In order to bridge the gap between the text driven image generation model and user needs, we propose an image generation and editing method named DiffBrush which is more in line with user drawing intuition - using ``Brush'' to generate paintings instead of ``Text''. With the assumption that the pretrained T2I model can successfully construct a one-to-one mapping from random noises to real images, we utilize the control of the latent denoising direction during the Latent Diffusion Model (LDM) denoising process to ultimately denoise a random noise and map it onto the image that the user needs. As shown in Fig.~\ref{fig:first_fig}, DiffBrush could paint on a generated or real image to realize image editing, and it could also paint on an empty canvas for controllable image generation. DiffBrush can also achieve more precise control by controlling the color, instance, and, semantics of specific pixel regions, as the last row of Fig.~\ref{fig:first_fig}.

Our specific contributions are as follows:

\begin{itemize}

\item We introduce a new framework named DiffBrush which utilizes the brush to achieve controllable generation and editing. Our DiffBrush is training-free, based on pretrained T2I models. Furthermore, it is almost compatible with all T2I models (stable diffusion(SD)~\cite{rombach2022high}, SDXL~\cite{podellsdxl}, Flux~\cite{flux}) that conform to thermodynamic diffusion processes, and it accepts the application of diverse Lora~\cite{hu2022lora} adjustment styles, rendering it an exceptionally user-friendly AI painting tool.

\item We propose two conditional guidance methods regarding corresponding generation targets, one for color appearance and the other for instance semantics, which to some extent solve the problem of how to design loss functions quantitatively in attention-based guidance methods.

\item We propose an initial noise refinement method, which takes the user sketch as the target and iteratively refines the initial randomly sampled noise to align it more closely to the distribution meeting user's need.

\end{itemize}

\section{Related Work}
\label{sec.Related work}

\subsection{Training-Based Image Generation.}
The rapid development of image generation is closely related to the T2I model. From SD~\cite{ramesh2022hierarchical}, which initially led the trend, to SDXL~\cite{podellsdxl}, which can generate high-resolution and high-quality images, to Lora~\cite{hu2022lora} and Pony~\cite{pony}, which are more distinctive, and Flux~\cite{flux}. On this basis, in order to lower the threshold for users to use and better control the generation results of images, researchers have made various improvements. ControlNet~\cite{zhang2023adding} adds additional encoders and cloning parameters to accept multiple modal control conditions, and ControlNext~\cite{peng2024controlnext} further improves based on this by reducing the number of parameters through a common VAE. 




\subsection{Training-Free Generation and Editing.} In addition to the training-based methods mentioned above, researchers have also provided many training-free, more user-friendly image generation and editing algorithms. We can simply divide it into three types. One is to control the generated image by providing image priori. SDEdit~\cite{meng2021sdedit} uses the diffusion model denoising mechanism to preserve a certain color condition in the generated image. In addition, researchers have proposed an editing algorithm based on attention. MasaCtrl~\cite{cao2023masactrl} achieves image editing by replacing and fusing instance features in self attention, PnP~\cite{hertzprompt} also achieves image editing by replacing, retaining, and adjusting weights in cross attention, and FPE~\cite{liu2024towards} analyzes the impact of two types of attention on generated images. There are also algorithms based on the characteristics of SDE~\cite{songscore} for editing. Self-guidance~\cite{epstein2023diffusion} achieves image editing by setting energy function targets in the cross-attention map, while FreeControl~\cite{mo2024freecontrol} achieves image editing of multimodal images by setting a similar feature library.



Training-based methods offer diverse controls but demand extra training due to additional modalities, causing costs to soar with model changes. Ordinary users seek cost-effective alternatives without iterations. Meanwhile, training-free methods, while cost-efficient, suffer from heavy reliance on reference images and imprecise control. In response to these issues, we introduce the DiffBrush framework. Leveraging pre-trained T2I models, DiffBrush enables users to paint, streamlining interaction, and enhancing controllability and accuracy in image generation.

\section{Methodology}

\subsection{Preliminary}
\textbf{Diffusion Sampling Process.}
The design of DiffBrush is based on the pretrained T2I models, all of which belong to conditional latent diffusion models~\cite{dhariwal2021diffusion}. Under the text prompt condition $c$, by training a temporal denoising module $\epsilon_\theta$, the randomly sampled standard normal distribution noise is gradually denoised and sampled into real image. Among them, $\epsilon_\theta$ usually chooses the Unet or DiT~\cite{Peebles2022DiT} structure, which is mainly composed of transformer blocks inside. These transformer blocks not only contain self-attention, but also can accept text prompt as the condition $c$ in cross attention blocks. The process is as follows:
\begin{equation}\label{eq1}
\begin{aligned}
\hat{\epsilon_t} &= \epsilon _\theta (z_t; t, c),\\
z_{t-1} &= Sample(z_t, t, \hat{\epsilon_t}, \alpha_i, \sigma_i),
\end{aligned}
\end{equation}
where $z_t$ is the random noise feature map at the timestep $t$, which has been encoded by the VAE encoder and denoised by $\epsilon_\theta$ $(T-t)$ times. $Sample()$ represents various diffusion sampling methods, such as DDPM~\cite{ho2020denoising}, DDIM~\cite{songdenoising}, etc. 


\noindent\textbf{Guidance.} 
According to Stochastic Differential Equations (SDE)~\cite{songscore}, the diffusion model actually belongs to score-based models, where the noise-perturbed score function represents the main direction of diffusion process, and $\epsilon_\theta$ can be seen as an approximate estimate of the score function of noise marginal distributions. Therefore, we can change the denoising sampling direction of the diffusion process by modifying the score function. Classifier guidance~\cite{dhariwal2021diffusion} is achieved by training an independent classifier to fit $p (c|x_t)$ for score based guidance. And classifier-free guidance~\cite{ho2022classifier} achieves similar results by adjusting the difference between conditional and unconditional predictions. The formulas as follow: 

\begin{equation}
\label{eq2}
\begin{aligned}
\hat{\epsilon}(x_t,c) &= \epsilon_\theta (x_t,c) -s\sigma_t\nabla_{x_t}\log{p(c|x_t)},  \\
\hat{\epsilon }(x_t, c) &= \epsilon _\theta (x_t, c) + s(\epsilon _\theta (x_t,c)-\epsilon _\theta (x_t)),
\end{aligned}
\end{equation}

In addition to class label, guidance can also be implemented by other conditions, such as a separate model for estimating energy function~\cite{liu2022compositional}, CLIP scores~\cite{nichol2022glide}, loss penalty from boundingbox in attention~\cite{chen2024training}, or targets and losses designed in attention~\cite{epstein2023diffusion,mo2024freecontrol}. These methods can be summarized in the following form:

\begin{equation}
\label{eq4}
\hat\epsilon_t = \epsilon _\theta (z_t; t, c) - s \sigma _t \nabla_{z_t} g(z_t;t,c),
\end{equation}
where $s$ denotes the hyperparameter weight to adjust the strength of the guidance, and $g(·)$ denotes the energy function designed for guidance.

\noindent\textbf{Distribution of latent space $Z$ encoded by VAE.}
\label{VAE_dis}
When training the Latent Diffusion Model (LDM)~\cite{ramesh2022hierarchical}, the VAE is trained separately. The real image $x$ is firstly encoded into $z$ by VAE encoder, and then decoded back into the real image $x'$ by decoder. By supervising $x$ and $x'$, the aligning between the encoder and decoder achieved. As mentioned in the paper~\cite{ramesh2022hierarchical}, the loss functions used in VAE are KL divergence loss function and MSE loss function. KL divergence loss is mainly used to control the distribution of latent feature space, while reconstruction loss is used to specifically compare the differences between $x$ and $x'$. 

\begin{figure}[t]
    \centering
    \small
    \includegraphics[width=1\linewidth]{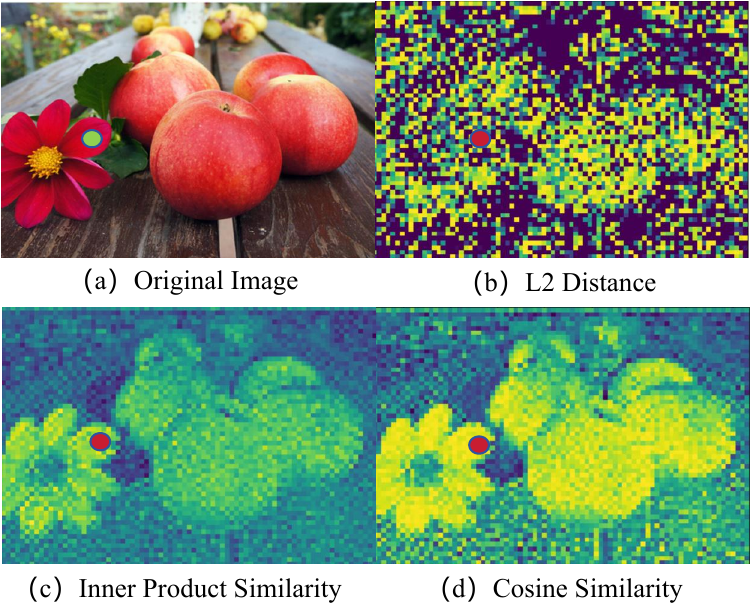}
    \vspace{-5mm}
    \caption{We selected a multi-instance image with similar colors but different semantics and encode it by VAE encoder into latent space. Since VAE uses MSE loss as the reconstruction loss for supervision, we first selected a pixel with similar color (a), calculated its MSE distance from all other pixel features in the latent space $Z$ (b), and then inner product similarity (c) and cosine similarity (d).}
    \label{fig:latent_visual}
\end{figure}

The two loss functions do not force the latent space transfer to a certain semantic space. So we visualize the distribution of latent space by different metrics to find out that which semantics is latent space related to. As shown in Fig.~\ref{fig:latent_visual}, we could find that the pixel features in latent space have very strong representational ability on color. The similar color pixels with different object class labels have high similarity even in different metrics. 

Therefore, we can reasonably infer that the latent space of stable diffusion is a feature space highly similar to the color space.


\begin{figure}[t]
    \centering
    \small
    \includegraphics[width=1\linewidth]{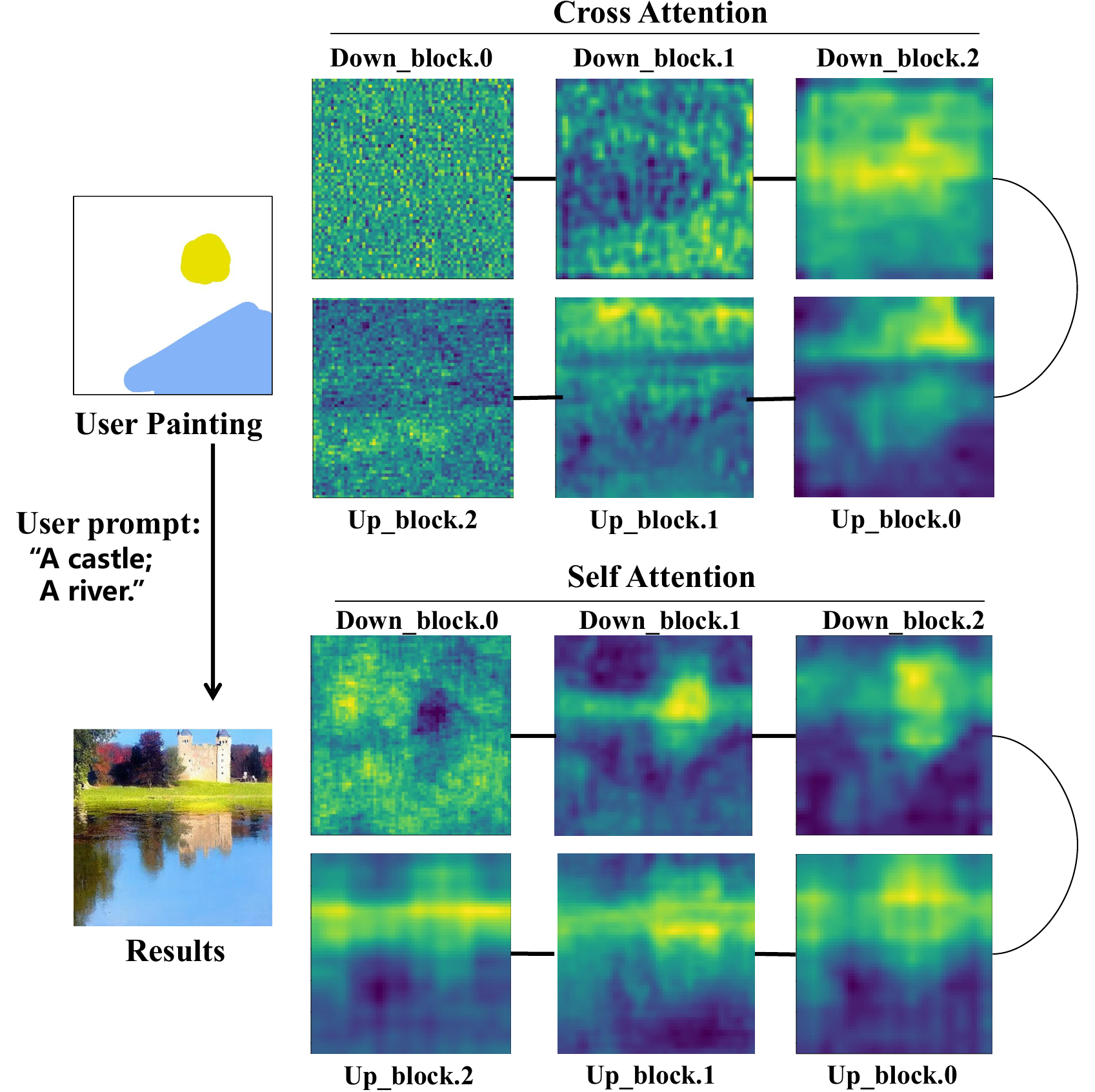}
    \vspace{-5mm}
    \caption{The visualization of the attention maps of different transformer layers in Unet of SD 1.5. We choose the cross attention map of "castle" and self attention map of its feature center to visualize. Furthermore, we could find the deeper layers like $Down\_block.2$ or $Up\_block.0$ have clearer instance or semantic directionality.}
    \label{fig:attn_visual}
\end{figure}

\noindent\textbf{Distribution of Attention Fitting in $\epsilon_\theta$.}
\label{attn_dis}
Although there is limited theoretical interpretability research on the denoising module in the diffusion model, researchers have observed certain statistical characteristics of the transformer module. Specifically, the denoising module widely incorporates cross-attention blocks pertaining to textual information. An interesting phenomenon is that as the layer goes deeper, the response expression of these cross-attention maps to textual information often becomes clearer, as shown in the Fig.~\ref{fig:attn_visual}. Based on this observation, we propose a hypothesis that there is a strong correlation between the semantic distribution of images generated by the diffusion model and the distribution of the deepest level cross-attention map. This means that if we change the value of the cross-attention map, the corresponding semantic objects and concepts in the image are likely to change accordingly. Similarly, for the self-attention map, we also assume that changing its value will change the position and state of the instance at the corresponding spatial position in the image.

\subsection{DiffBrush Framework}
\begin{figure*}[]
    \centering
    \includegraphics[width=2.0\columnwidth]{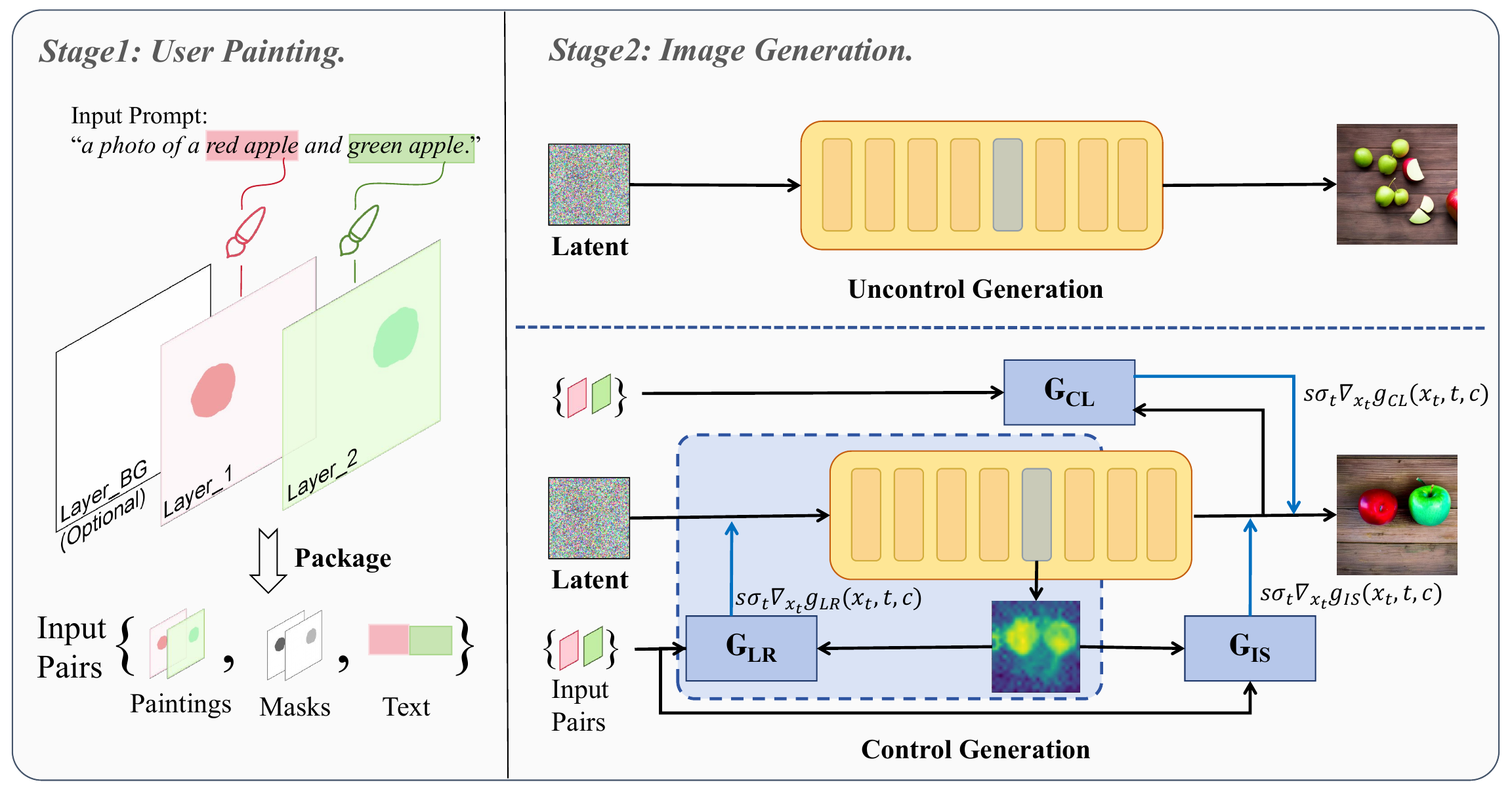}
    \vspace{-3mm}
    \caption{DiffBrush framework comprises two stages: user painting and image generation. In the user painting stage, user inputs text, selects instances and attributes as brush semantics, draws on canvas (with different instances on separate layers), and also can edit based on a reference image. The result, mask, and semantics are packed into a triplet. In image generation, DiffBrush uses color, instance, and starting point constraints to guide generation, which is compatible with multiple models, and employs energy functions to balance user conditions and model freedom for generating desired images.
    }    
    \label{fig.Method}
\end{figure*}

Based on the above premises and assumptions, we propose DiffBrush which is mainly divided into two stages as shown in the Fig~.\ref{fig.Method}. The first stage is the user painting stage. The user initially inputs a textual description of the entire painting, then selects the desired instances and their corresponding attributes to be painted and edited within the prompt, regarding them as the semantics labels of brushes to paint on the canvas. There is no need to draw details, only to ensure that the color and scale shape are roughly correct. DiffBrush can complete the details based on semantics in the generation stage. Different instances need to be painted on different layers to ensure independence between instances, avoiding color confusion and instance fusion. Additionally, users have the capability to perform image editing by utilizing existing reference images as background layers and drawing based on them. DiffBrush covers the drawing content to the corresponding position of the reference image according to text prompt while ensuring that the overall background image remains unchanged, and ensures image harmony. At the end of the user session, DiffBrush will package the user's drawing results, the corresponding mask, and the semantics of each brush into a triplet tuple for image generation in subsequent stages.

In the second stage, DiffBrush begins to guide image generation based on user input triplets. The image generation process of DiffBrush is based on T2I models, compatible with the series of SD,  SDXL, and Flux, etc., and does not require additional training, belonging to training-free guidance. Similar to Self-Guidance~\cite{epstein2023diffusion} and FreeControl~\cite{mo2024freecontrol}, DiffBrush is also designed based on the Langevin Dynamics Sampling~\cite{chan2024tutorialdiffusionmodelsimaging} for guidance. But the difference is that Self-Guidance and FreeControl rely on real reference images provided by the user during guidance, and achieve image editing by manipulating the attention map responded to by instances in the real image in the attention block, while there is no real images or real instances to refer for DiffBrush, only rough hand drawn images without textures provided by users. 

How to balance the strength of the conditions drawn by users and the freedom of image generation of the model automatically is the problem that DiffBrush needs to solve. Facing with this challenge, we have designed three energy functions to guide the T2I models, namely color guidance~(CL), instance\&semantic guidance~(IS), and latent regeneration~(LR), working from the perspectives of color, instance semantics, and distribution. These guidances work independently for each instance in the generated image, as shown in the Fig~.\ref{fig.Method}. The complete formula is set as follows:

\begin{equation}
    \label{eqsum}
    \begin{aligned}
    z_T &= z_T + \sum G_{LR}, \\
    \hat\epsilon_t &= \epsilon _\theta (z_t; t, c) + G_{CL} + G_{IS},
    \end{aligned}
\end{equation}
where $G_{LR}$ is for latent regeneration, $G_{CL}$ is for color guidance, and $G_{IS}$ is for instance \& semantic guidance. For details, please refer to the following subsection.

\subsection{Color Guidance}

As we know from section~\ref{VAE_dis}, the similar RGB pixel values in real images will inevitably have similar mapping features in the latent space. Although VAE achieves a certain degree of perceptual information compression and semantic information extraction, the overall features are still biased towards the color space, as shown in the visualization Fig.~\ref{fig:latent_visual}.

From the Fig.~\ref{fig:latent_visual}, it can be observed that whether measured by Euclidean distance, inner product similarity, or cosine similarity, the color information of features in latent space $Z$ exist in an explicit form, and similar colors have similar distances in the above three metric spaces. From this, it can be seen that in order to achieve color control, it is only necessary to move the potential pixel features of the painting target towards the corresponding color features in the above three metric spaces. Inspired by this, we designed an energy function corresponding to color control for guidance. The formula is as follows:

\begin{equation}
\label{eq6}
\begin{aligned}
G_{CL} & = s_{cl}\sigma _t\nabla_{z_t}g_{CL}(z_t, z^p_{t})\\
 & = s_{cl}\sigma _t\nabla_{z_t}MSE(z_t, z^p_{t})\\
 & = s_{cl}\sigma _t\nabla_{z_t} (z_{t}-z^p_{t})^2\\
& = 2 s_{cl}\sigma _t(z_{t}-z^p_{t}),
\end{aligned}
\end{equation}
where $s_{cl}$ is a hyperparameter to adjust the intensity of color control, and $g_{CL}()$ is an energy function designed for color control guidance, which can be any loss function based on Euclidean space, inner product space, or cosine distance space. Here, we chose the same MSE loss function in VAE training as the energy function. $z_t$ is the latent feature with timestep $t$, and $z_t^p$ is the user drawing feature that has been encoded in encoder and denoised by diffusion process.

\subsection{Instance \& Semantic Guidance}
\label{G_{is}}

Although the performance of color control guidance is quite good, it can basically control the color of the corresponding pixels in the generated image and the position distribution of instances. But there are still problems that color control cannot solve, such as, difficulty in distinguishing between similar color instances, and confusion in assigning semantic attributes to text.

\begin{figure}
    \centering
    \includegraphics[width=1\linewidth]{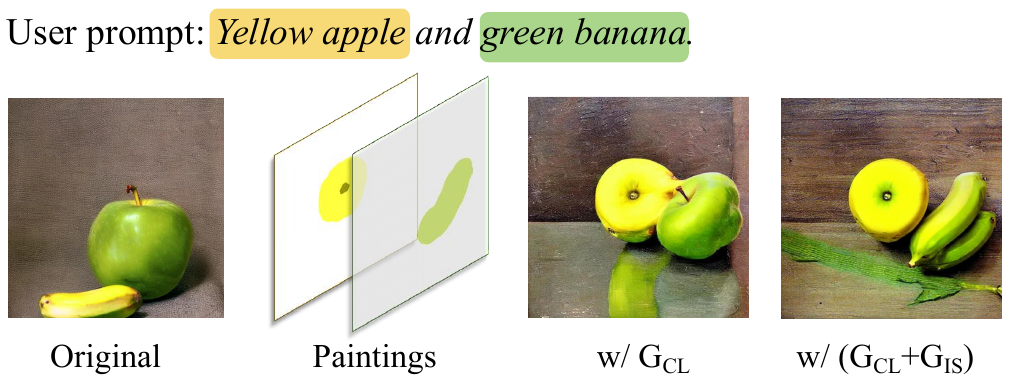}
    \vspace{-5mm}
    \caption{The effect of the Ins\&Sem guidance. It shows that the original output under prompt condition, user's painting, the output with $G_{CL}$, and the output with $G_{CL}$ and $G_{IS}$.}
    \label{fig:IS}
\end{figure}

In the T2I model, the text attributes and concepts of instances have strong tendencies, which are related to the training dataset and the distribution of instances in the real world. For example, as shown in Fig.~\ref{fig:IS}, generally speaking, when the concept of an apple is mentioned, its color attributes tend to be green, red, etc.; when the concept of a banana is mentioned, its color attributes tend to be yellow. When no additional control conditions are applied and only ``yellow apple and green banana'' are input as text prompts, although green and yellow are color - attribute modifiers of each other respectively, since the instance itself has a color tendency in semantics, even if its own color - attribute modifier modifies itself, the instance still retains a certain mainstream color - attribute tendency in terms of color attributes. In addition, since the text encoder is of the transformer structure, the instance semantics are affected by the features of all other tokens in the whole sentence during encoding, which further reinforces the mainstream color tendencies of ``green apples'' and ``yellow bananas'', resulting in the misalignment of color attributes in the original image.

To address this issue, we designed Instance \& Semantic Guidance, which applies guidance similar to color control on semantics and instances. In order to find energy functions related to semantics and instances, we referred to Self Guidance~\cite{epstein2023diffusion}, FreeControl~\cite{mo2024freecontrol} and other methods, and used self-attention map and cross-attention map as explicit representations of instances and semantics to design energy functions. As shown in Fig.~\ref{fig:attn_visual}, taking SD1.5 as an example, we can see that in the Unet downsampling and upsampling modules, the cross-attention map and self-attention map indeed pay more attention to instances and semantics, and it is appropriate to use them as the object for guidance.

However, designing what kind of energy function is still a problem. Self-Guidance and FreeControl rely on instances in the reference image to provide their value distribution in the attention map, ensuring instance consistency. However, DiffBrush's reference image is a user-drawn sketch that contains almost no texture or semantic information. How to set its value distribution target in the attention map is the problem that DiffBrush needs to solve.

\begin{algorithm}[h]
\caption{Attention-based Ins\&Sem Guidance}
\label{insguidance}
\KwData{\\ 
Ins center index $i_{ins}$, Sem token id $i_{sem}$ ,\\ 
Ins\&Sem Masks $M_{IS}$,  \\
self attention map $A_{self} \in R^{HW \times HW}$, \\
cross attention map $A_{cross} \in R^{HW \times T}$,\\
Loss function $L_{p}(X, a, M), L_{n}(X, a, M)$,\\
loss weight $\lambda_{p}, \lambda_{n}$,
}
\KwResult{Ins\&Sem energy function $g_{IS}$,}
$g_{IS} \leftarrow 0$ \;
\textbf{Define}: \\
$L_{p}(X, a, M) = AVG(-\exp{(X - a)*M}-1 )$\;
$L_{n}(X, a, M) = AVG(\exp{(X - a)*M}-1 )$\;
\For{$M_{ins} \in M_{IS}$}{
    $A_{ins}\in R^{H \times W}\leftarrow A_{self}[:, i_{ins}]$\;
    $A_{sem}\in R^{H \times W}\leftarrow A_{cross}[:, i_{sem}]$\;
    \For{$A \in [A_{ins}, A_{sem}]$}{
        $A_{p}, A_{n} \leftarrow A[M_{ins}], A[\sim M_{ins}]$\;
        $mean_{p} \leftarrow AVG(A_{p})$\;
        $mean_{n} \leftarrow AVG(A_{n})$\;
        $M_{p} \leftarrow A_{p} \leq mean_{p} $\;
        $M_{n} \leftarrow A_{n} \geq mean_{n} $\;
        $g_{IS} \leftarrow g_{IS} + \lambda_{p} L_{p}(A_{p}, mean_{p}, M_{p})$\;
        $g_{IS} \leftarrow g_{IS} + \lambda_{n} L_{n}(A_{n}, mean_{n}, M_{n})$\;
    }
}
\end{algorithm}

Consequently, we design a instance-level distribution-based guidance method. Since we cannot obtain precise pixel-level guidance groundtruth, we have decided to start with the overall distribution of instances. The sketches drawn by users can provide masks corresponding to instances. Pixel features belonging to the same instance must have high attention correlation in the self-attention map. Therefore, we take the feature closest to all other features on the average distance in the mask as the instance feature center and select its corresponding self-attention map as the instance-related supervision target. As for semantics, since we can obtain the semantics of the brush corresponding to the masks, so we can also obtain the cross-attention map corresponding to the instance token, which is also used as a semantic-related supervision object. The detailed algorithm is shown in Alg.~\ref{insguidance}.

The core idea of Alg.~\ref{insguidance} is to utilize masks to adjust the features of instances in the self-attention map and cross-attention map. Specifically, for the internal feature of the instance, when the attention value corresponding to the feature is smaller than the internal attention mean, it will be brought closer to the mean direction to achieve the enhancement effect on the weak feature. For external features of an instance, if their corresponding attention value is greater than the overall attention mean, it will also move towards the mean direction, thereby achieving the goal of weakening such strong features.
Ultimately, guidance will still be implemented in the form of Equ.~\ref{eq4}, as follows:
\begin{equation}
    \label{7}
    G_{IS} = s_{is}\sigma_t\nabla_{z_t}g_{IS}(z_{t}, \epsilon_\theta,t,M, \lambda),
\end{equation}
where $M$ is the mask corresponding to the user's sketch. For the calculation of $g_{IS}$, please refer to Alg.~\ref{insguidance}.

\begin{figure}[t]
    \centering
    \includegraphics[width=1\linewidth]{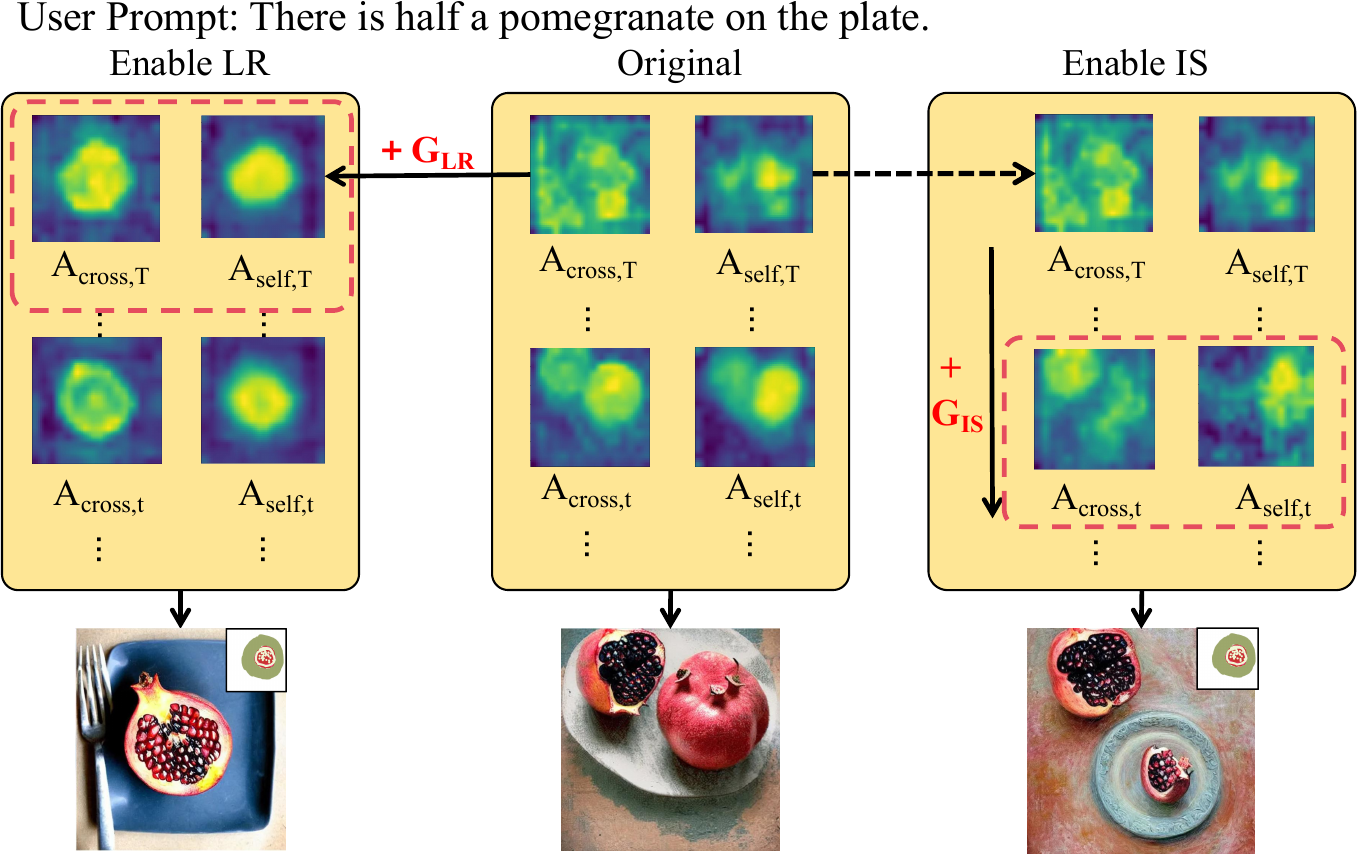}
    \vspace{-3mm}
    \caption{Visualization of the changes in the attention maps under different guidance.}
    \label{fig:attentionchanges}
\end{figure}

\subsection{Latent Regeneration}

Color control and instance \& semantic control make it possible for users to accurately control instances in the generated images. However, in the actual generation process, the random noise sampled in initialization still has a significant impact on the generated results. In the whole noise distribution, some random seed samples are more in line with the instance distribution of the user's drawing, resulting in higher image quality and better meeting the user's needs. Therefore, we also designed a scheme called Latent Regeneration (LR) to find initial noise that matches the distribution of user sketches. Similar to Ins\&Sem control, LR also utilizes the explicit representation feature of the T2I model on the attention map. In the first step $t=T$, $G_{LR}$ is circularly calculated and the gradient is superimposed on $Z_T$ to gradually update the latent with smaller hyperparameter weights, achieving refinement of the initial noise. And the initial noise latent would be transfered a new distribution which is more suitable for user paintings. The formula is set as follows:

\begin{equation}
    \label{eq8}
    G_{LR} = \lambda_{LR}\sigma_T\nabla_{z_T}{g_{LR}(z_{T},t,\epsilon_{\theta}, M)},
\end{equation}
where the calculation of $g_{LR}$ is the same as $g_{IS}$ as shown in Alg.~\ref{insguidance}. Under the LR mechanism, the initial noise distribution changes, and the final generated result also changes accordingly. About the difference between $G_{LR}$ and $G_{IS}$, as shown in Fig.~\ref{fig:attentionchanges}, the $G_{LR}$ operates on the first step, and the attention map has been significantly changed like the user paintings. While $G_{IS}$ operates during the denoising process, the attention map changes softly.


\section{Experiment Results}

\subsection{Experiment Setup}
\noindent\textbf{Evaluation.} To ensure a certain level of fairness, we referred to the experimental setup of FreeControl~\cite{mo2024freecontrol} and selected the ImageNet-R-TI2I~\cite{tumanyan2023plug} dataset for evaluation. This dataset contains 30 images from 10 categories, each with five text prompts for text guided image to image translation. It is worth noting that due to different input conditions, other methods such as FreeControl use input conditions such as Canny or Sketch, \emph{etc.}, while DiffBrush gets stroke as input condition. For evaluation, we also choose CLIP score~\cite{radford2021clipscore} and LPIPS distance~\cite{zhang2018unreasonable} as metrics. The CLIP score can be used to characterize the degree of matching between text and images. LPIPS calculates the semantic and structural similarity between two images. 

\noindent\textbf{Implement Details.} DiffBrush is developed based on the PyTorch framework and is compatible with SD series, SDXL, and Flux models. The results in the main text are generated based on SD1.5. DiffBrush has low hardware requirements and can run smoothly on Nvidia RTX3090. The hyperparameters related to SD 1.5 use default values. The default hyperparameters for DiffBrush are: $s_{cl}=5, s_{is}=1, \lambda_{LR}=0.1$. The above hyperparameters will be modified according to the different text and drawing. In the selection of attention maps, we chose $up.block.0.attn2$ and $down.block.1.attn1$. $G_{CL}$ and $G_{IS}$ works in the 0-50\% stage of the inference process, with some fluctuations depending on the prompt and painting. More details please refer to our supplementary materials.

\subsection{Quantitative and Quality Results}

To ensure fairness, we replicated the baseline experiment of FreeControl and drew 150 strokes corresponding to text prompts for 30 images in ImageNet-R-TI2I. We also conducted supplementary testing on various baseline algorithms. And the optimal results of each method under various conditions were selected for comparison. Please refer to Tab.~\Ref{tab.results} for specific results. From the table, it can be seen that DiffBrush achieved better results than other methods under the condition of user drawing. It is worth mentioning that SDEdit obtained better results than before under the Canny condition after inputting the stroke condition.

\begin{table}[h]
\renewcommand\arraystretch{1.3}
\centering
\setlength{\tabcolsep}{2mm}
\caption{Quantitative results. SE represents SDEdit-0.75~\cite{meng2021sdedit}, and SE* represents SDEdit-0.85, PNP represents ~\cite{tumanyan2023plug}, FC represents ~\cite{mo2024freecontrol}. The second, third, and fourth lines represent the best results achieved by each model under their respective optimal conditions.}
\label{tab.results}
\small
\begin{tabular}{lcccccc}
\hline
Method    & SE & SE* & P2P   & PNP    & FC & DiffBrush   \\ \hline
Cond & Stroke       & Stroke       & HED   & Normal & Canny       & Stroke \\ \hline
CLIP↑     & 0.302       & 0.317       & 0.253 & 0.286  & 0.322       &  0.326      \\
LPIPS↑    & 0.547       & 0.710       & 0.194 & 0.347  &  0.724       &  0.738      \\ \hline
\end{tabular}
\end{table}

Similarly, we also provide qualitative analysis results. As shown in Fig.~\ref{fig.results}, we provide excellent performance of DiffBrush in different scene requirements. Even when Lora with different styles such as oil painting and traditional Chinese painting is loaded, DiffBrush can still achieve strict control over color and instance semantics. Latent regeneration also reduces the deviation that may occur in color control and Ins\&Sem control, making the generated images closer to the style that users need. Compared with SDEdit~\cite{meng2021sdedit}, DiffBrush can fully utilize the color and instance information provided by stroke, and perform better in terms of instances, semantics, and textures. In addition, due to the lack of suitable reference images, Self-Guidance~\cite{epstein2023diffusion} and FreeControl~\cite{mo2024freecontrol} are unable to make specific edits to the targets in the images, resulting in images that do not meet the requirements.

We provide addition user study in the supplementary materials.

\subsection{Ablative Study}

Quantitatively determining the efficacy of guidance in image controllable generation poses challenges; however, as depicted in Fig.~\ref{fig.results}, it is evident that each proposed guidance exerts an influence on the original image and adheres to certain statistical patterns. After adding $G_ {CL}$, the pixel color corresponding to the painting position is significantly constrained, but lacks semantic constraints, which can easily lose style and reality. $G_ {IS}$ focuses more on maintaining the correctness of instance semantics in the image, but the effect of using it alone is not good. Enabling it together with $G_ {CL}$ can achieve better guidance effect, and the correctness of image color, instance, and semantics can be guaranteed. The addition of $G_ {LR}$ further optimized above effect, resulting in a significant change in the layout of the image and a more harmonious overall effect. 

We also provide addition ablative study in the supplementary materials.

\section{Conclusion}
In this paper, we propose a training-free controllable image generation method named DiffBrush based on the T2I model, which accepts user hand-drawn control. Compared with other controllable image generation methods, DiffBrush not only eliminates additional training costs, but also controls semantics on the basis of color control. Compared with image editing methods, DiffBrush provides a new guidance solution with instance-level masks, solving the problem of inaccurate editing of instances without reference targets. At the same time, it achieves color control matching with latent space in the form of guidance. Although DiffBrush can achieve good guidance effects in color, instance, and semantics, the strength of guidance conditions still needs to be adjusted by users themselves, which is not only a pain point for user operation, but also an improvement direction for our future work.

%
%
%
%

\bibliographystyle{ACM-Reference-Format}
\bibliography{main}


\appendix

\newpage
\begin{figure*}[t]
    \centering
    \includegraphics[width=0.975\linewidth]{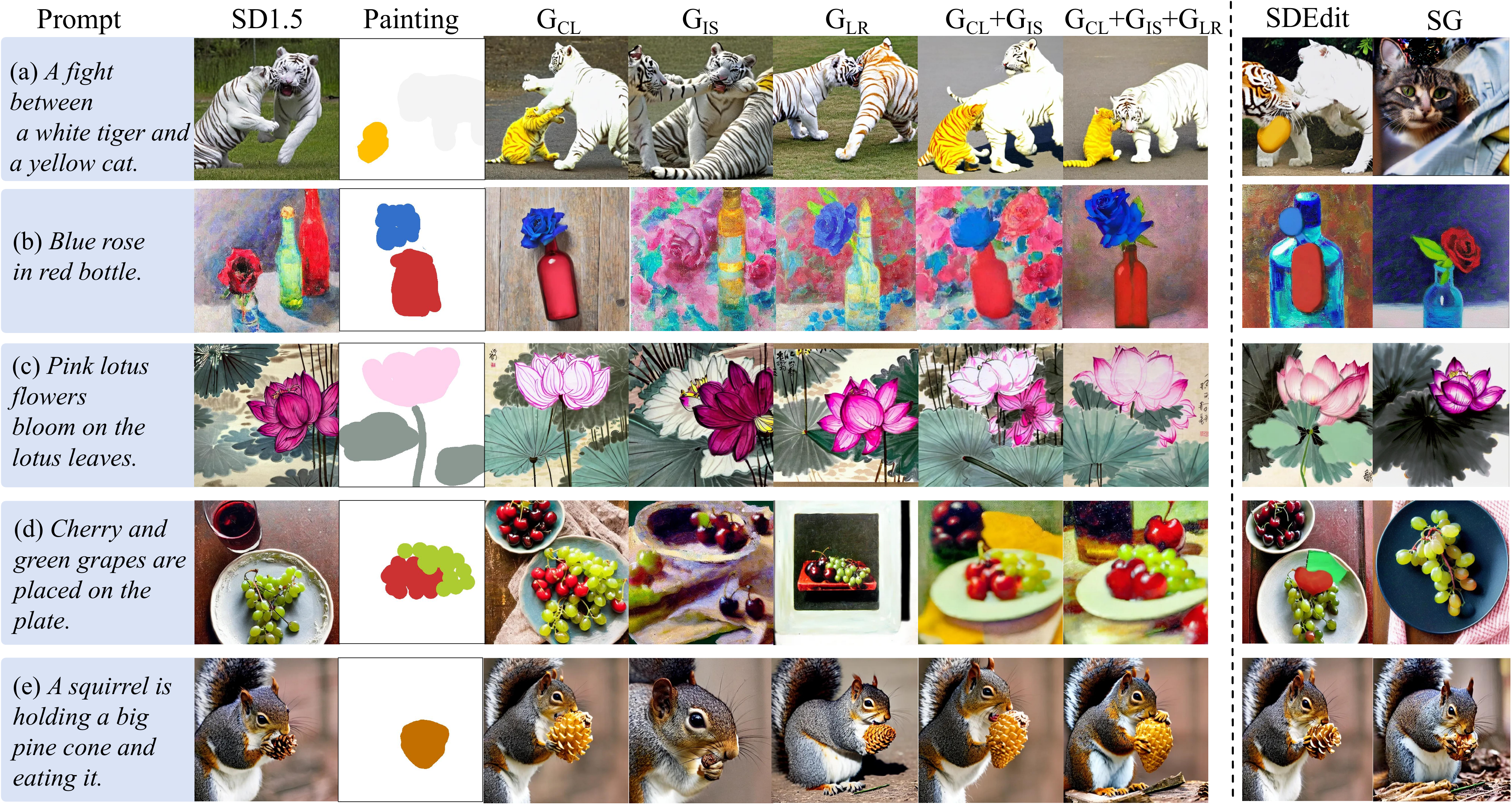}
    \caption{Qualitative results of DiffBrush. We provide visualization results of DiffBrush under different Lora. (a) None, (b) oil-painting, (c) Chinese painting, (d) oil-painting-2, as well as the impact of different guidance combinations on image generation. There are also comparisons with the classic stroke-based method SDEdit~\cite{meng2021sdedit} and image editing method Self-Guidance~\cite{epstein2023diffusion}.We also provide more visualization content in the appendix, including DiffBrush effects based on other T2I models.}\label{fig.results}
\end{figure*}

\begin{figure*}[h]
    \centering
    \includegraphics[width=0.975\linewidth]{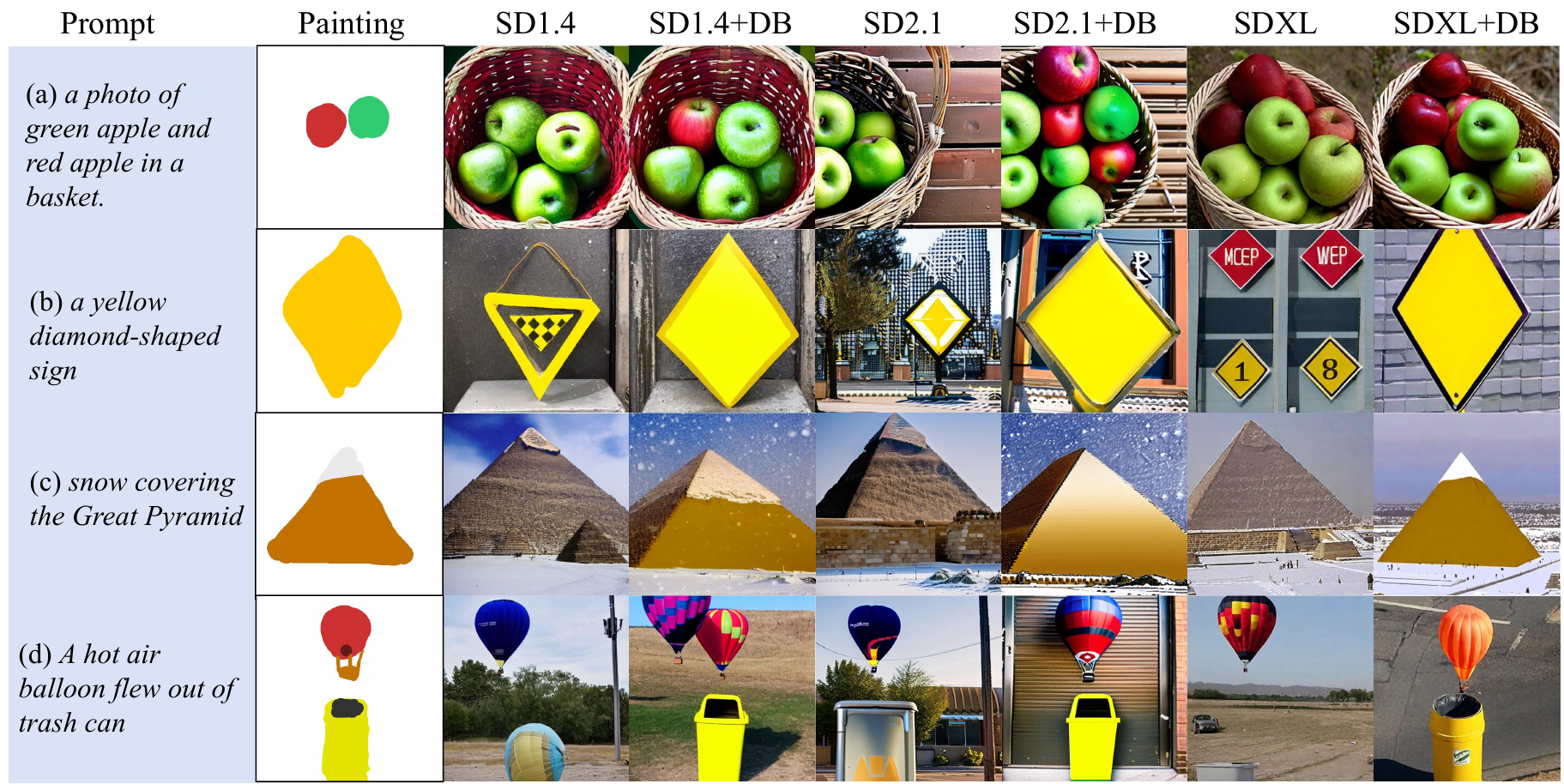}
    \caption{We provide text-generated images of different T2I models under the same prompt and with the assistance of DiffBrush (DB). In addition, we plan to update the visualization and evaluation related to Flux+DiffBrush in future versions.}
    \label{fig:add_results}
\end{figure*}

\newpage

\begin{figure*}[t]
    \centering
    \includegraphics[width=0.915\linewidth]{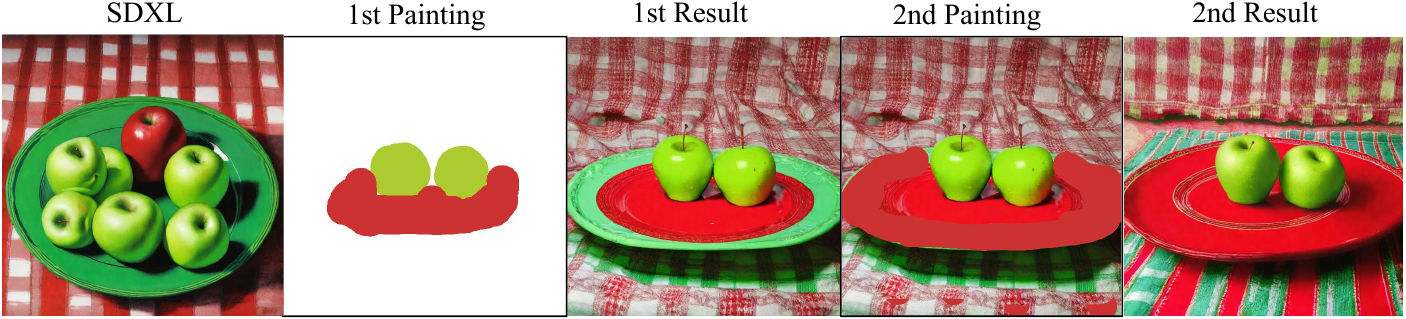}
    \caption{The prompt is "Green apples on the red plate". Based on SDXL 1.0 model, DiffBrush also could generate the high-quality images with $1024^2$ resolution. In addition, by inputting the first result as the background, DiffBrush allows users to edit the image by repainting.}
    \label{fig:Imageedit}
\end{figure*}

\begin{figure*}[h]
    \centering
    \includegraphics[width=0.915\linewidth]{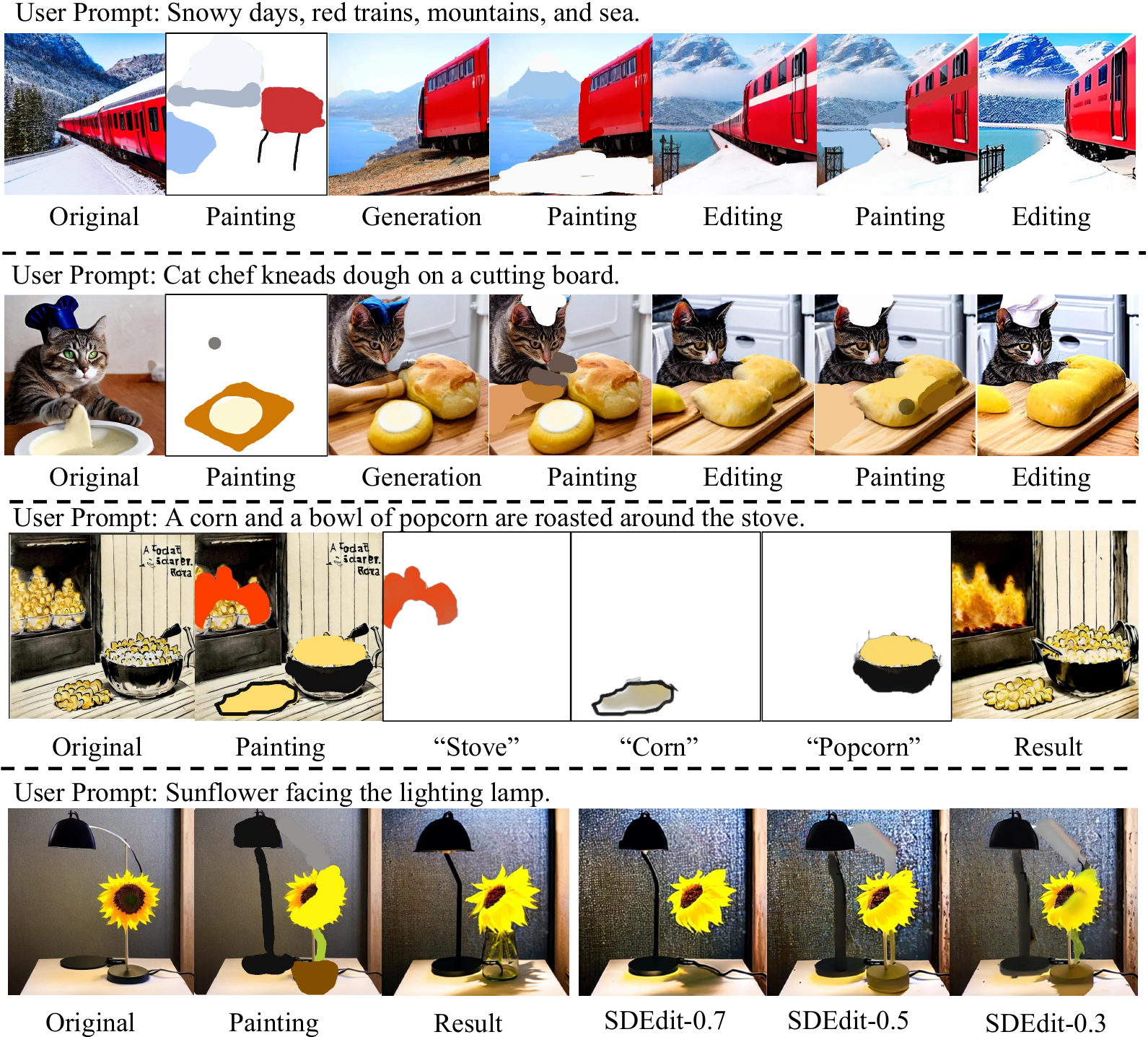}
    \caption{Based on SD1.5, DiffBrush also could realize image editing by multiple times painting. On the first row, we show the elements used in editing process. DiffBrush need user to paint the instance object in different layer with their semantic labels. And on the second row, we show another editing result and the comparison with SDEdit~\cite{meng2021sdedit} under the same painting and different strength.}
    \label{fig:Editresult}
\end{figure*}

\newpage
\appendix
\section{More implementation details}
\subsection{Compatibility with Other T2I Methods}
\label{sec:Compatibility}

As a training-free guidance, DiffBrush has plug-and-play capability and is suitable for almost all image generation methods that conform to the diffusion process. However, the settings of relevant parameters may vary for different models. In this section, we demonstrate the compatibility of DiffBrush on SD1.4, SD2.1, and SDXL 1.0.

We chose the 512$^2$ resolution version of SDXL 1.0, which has been finetuned by users, because the high computational cost of gradient backward with a native resolution of 1024$^2$ poses a challenge to the memory size of commonly used consumer grade graphics cards. But this does not mean that DiffBrush cannot be applied to larger resolutions.

In addition, during the sampling process, we employ the PNDM scheduler~\cite{liu2022pseudo} as our default choice.

\subsection{Config of DiffBrush}

In the context of DiffBrush, an extensive array of hyperparameters is available for adjustment. Each instance layer within the user paintings is equipped with its distinct guidance hyperparameter. A detailed enumeration will be carried out in accordance with different categories of guidance:

\noindent \textbf{Color Guidance.} Regarding the color guidance of each instance layer, the hyperparameters consist of the guidance period and the guidance strength$s_{cl}$. The former governs the timesteps over which the guidance influences, while the latter determines the intensity of such influence. It is feasible to assign different strengths corresponding to diverse periods. Moreover, the color guidance encompasses a background layer, which also possesses the identical hyperparameters of period and strength. These hyperparameters are employed to regulate the intensity of the impact on the background image. The range of guidance period is flexible, mainly set to 0\% - 50\%. The default set is from 0\% to 25\%, $s_{cl}$ is about 5, the background strength is about 1.5.

\noindent \textbf{Ins\&Sem Guidance.} Because Ins\&sem guidance is related to text vectors and tokenized tokens, there are more hyperparameters. Each layer has to be linked to its corresponding text tokens. When setting up the guidance, the guidance period and guidance strength for each token need to be configured. Generally, tokens from the same instance or layer can be set to the same value. But in some special situations, for example, if there are attribute tokens that need extra strengthening among the tokens, we should increase their strength separately.
In addition to the strength related to the text, there is also the strength related to the instance pixels. This is used to enhance the integrity and boundaries of the instance. The effective range of the guidance period is flexible, ranging from 0\% to 70\%, with a default of 0 - 25\%. Since it impacts attention, the strength setting is non-linear, and values of the strength range from 0 to 300 may all be effective.

\noindent \textbf{Latent Generation.} The hyperparameters of Latent Regeneration are similar to those of Ins\&Sem Guidance, both involving hyperparameter settings related to attention. The difference lies in the scope of influence. Latent Regeneration only affects the denoising process in the initial first step and repeats multiple times. Therefore, there is an additional hyperparameter N for setting the number of repetitions. Generally, the default value of N is 10.

\subsection{How to get the center of feature in self-attention block?}
In Sec.~\ref{G_{is}}, we introduce the guidance of instance and semantic. But in the manuscript, we ignore the method how to find the feature center, as named as instance center index $i_{ins}$ in the Alg.~\ref{insguidance}. So we supply the details of the function here.

\begin{algorithm}[h]
\caption{Attention-based Ins\&Sem Guidance}
\label{ins_index}
\KwData{\\ 
Ins\&Sem Masks $M_{IS}$,  \\
self attention map $A_{self} \in R^{HW \times HW}$, \\
cross attention map $A_{cross} \in R^{HW \times T}$,\\
}
\For{$M_{ins} \in M_{IS}$}{
    $A_{ins,pos}\in R^{n \times n}\leftarrow A_{self}[Mask, Mask]$\;
    $A_{ins,neg}\in R^{n \times HW-n}\leftarrow A_{self}[Mask, ~Mask]$\;
    $mean_p$ = $AVG(A_{ins,pos},dim=1)$\;
    $mean_n$ = $AVG(A_{ins,neg},dim=1)$\;
    $Diff$ = $mean_p - mean_n$\;
    $i_{argmax}$ = $Diff.argmax()$\;
    $i_{ins}$ = $where(M_{ins})[i_{argmax}]$\;
}
\end{algorithm}

\section{Operational Efficiency Analysis}

As an AI drawing tool prioritizing user-friendliness, our objective is to achieve cost-effective performance on consumer-grade graphics cards. Therefore, we conducted speed and memory tests on DiffBrush. The results are shown in the Tab.~\ref{tab.runtime}.

\begin{table}[h]
\renewcommand\arraystretch{1.3}
\centering
\setlength{\tabcolsep}{3mm}
\caption{The spatial and temporal analysis during inference of the DiffBrush on different T2I models. Even if the SDXL model is selected, with all guidance enabled at $512^2$ resolution, the maximum memory usage still does not exceed 24GB, and it can run successfully on consumer grade graphics cards.}
\label{tab.runtime}
\small
\begin{tabular}{lcccc}
\hline
 & SD1.4  & ...+$G_{CL}$ & ...+$G_{IS}$ & ...+$G_{LR}$ \\\hline
Time (s)   & 1.82  & 2.32       & 2.98             & 4.03                   \\
Mem (MiB)    & 3999 & 4035      & 7365            & 7365                  \\ \hline
 & SDXL   & ...+$G_{CL}$ & ...+$G_{IS}$ & ...+$G_{LR}$  \\\hline
Time (s)   & 3.42  & 4.41       & 9.15             & 12.6                   \\
Mem (MiB)    & 9373 & 9431      & 23481           & 23481                 \\ \hline
\end{tabular}
\end{table}

The spatial and temporal evaluation were deployed on a RTX 4090D, generating images with a resolution of $512 ^ 2$, and the model was set as float16 type. As shown in the Tab.~\ref{tab.runtime}, even with all guidance enabled, DiffBrush still does not exceed the maximum video memory of a consumer grade graphics card.

\section{Different attention layers for guidance}
We also provide comparative results obtained by guiding on different attention blocks, partially visualized as follows:

\begin{figure*}
    \centering
    \includegraphics[width=1\linewidth]{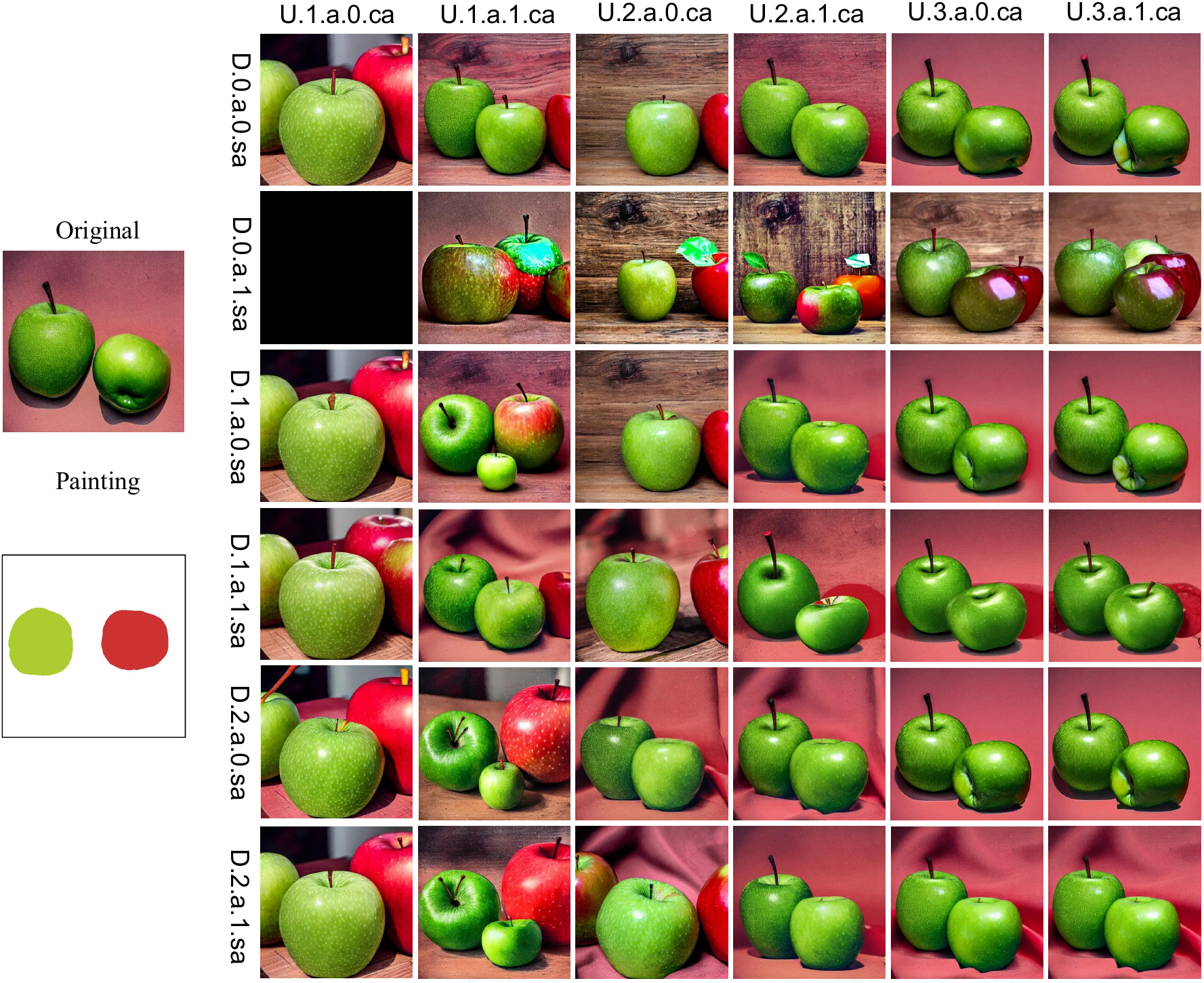}
    \caption{Prompt is: "a photo of a green apple and a red apple." Only Ins\&Sem guidance influence different transformer blocks. "U.1.a.1.ca" represents "up\_blocks.1.attentions.1.transformer\_blocks.0.attn2", "D.2.a.1.sa" represents "down\_blocks.2.attentions.1.transformer\_blocks.0.attn1"}
    \label{fig:trans}
\end{figure*}

\section{User study}

We recruited 20 unrelated volunteers to participate in the user study for comparison between DiffBrush and the traditional controllable image generation methods. We ask volunteers imagine a painting and write its concise text description. Then we provide them with a web canvas to draw a simple coarse painting for their imagination. And we start to conduct two different experiment for controllable generation and editing. We choose SDEdit and Self-guidance as baseline, and make simple UIs for them based on gradio. We generate reference images by text descriptions based on the T2I model in advance for editing pipeline.

For controllable image generation, we require volunteers input their text prompt and painting into the DiffBrush demo and SDEdit demo, generating images that match their imagination as much as possible within 10 minutes. And then they will score the generated images from SDEdit and DiffBrush. We record their comments, scores and time cost.

For editing pipeline, we provide the volunteers with a basic image, requiring them to edit the image by DiffBrush, SDEdit, and Self-Guidance to ideal status as much as possible within 10 minutes. Same as before, the volunteers will score their final paintings, we will record them. And the results as follow:

\begin{table}[t]
\centering
\caption{\small{Comparison between DiffBrush, SDEdit and Self-Guidance. "CIG" represents controllable image generation. "DB" represents DiffBrush; "SG" represents Self-Guidance.}}
\small
\renewcommand\arraystretch{1.2}
\setlength{\tabcolsep}{2mm}
\begin{tabular}{lccccccc}
\toprule
                & DB-CIG & SDEdit & DB-Edit &SDEdit & SG \\ \midrule 
Time cost~(min)$\downarrow$ & 9.3   & 8.5   & 9.9   & 5.6  & 9.9  \\
Score~($0-10$)$\uparrow$ & 8.25 & 7.75  & 8.50  &  8.00   &  7.25   \\
\bottomrule
\end{tabular}\label{tab:userstudy}
\end{table}

We could find that the time cost of DiffBrush is longer than SDEdit. This is because most volunteers have been finely adjusting hyperparameters and paintings to approach the ideal state, while SDEdit takes less time because the adjustable content and direction are limited, and the final generated image quality is also limited. Self Guidance did not perform well in the user study, and volunteers could not feel and use the demo well because its usage was slightly abstract and not intuitive.

\section{Failure case}

Although DiffBrush looks powerful, there is still a lot of room for improvement compared to the ideal painting tool. As DiffBrush is a Training Free method, the generated images still rely on the basic capabilities of pre trained T2I models. The stronger the capabilities and the more data learned by the model, the more ideal the images generated and edited by DiffBrush will be. Faced with some targets and backgrounds that the model itself is not good at, even with guidance in color, instance, and semantics, DiffBrush cannot produce images that meet the requirements.

In addition, DiffBrush is not proficient in achieving complex textures. This issue can be attributed to two aspects. On one hand, the pre - trained T2I model itself has limited capabilities in this area. On the other hand, it is extremely challenging to strike a balance between the sketch conditions with rough textures and complex, fine, and realistic images. It is necessary to repeatedly adjust the relevant hyperparameters to obtain satisfactory results.

We also provide some visualization of failure cases:

\begin{figure*}
    \centering
    \includegraphics[width=1\linewidth]{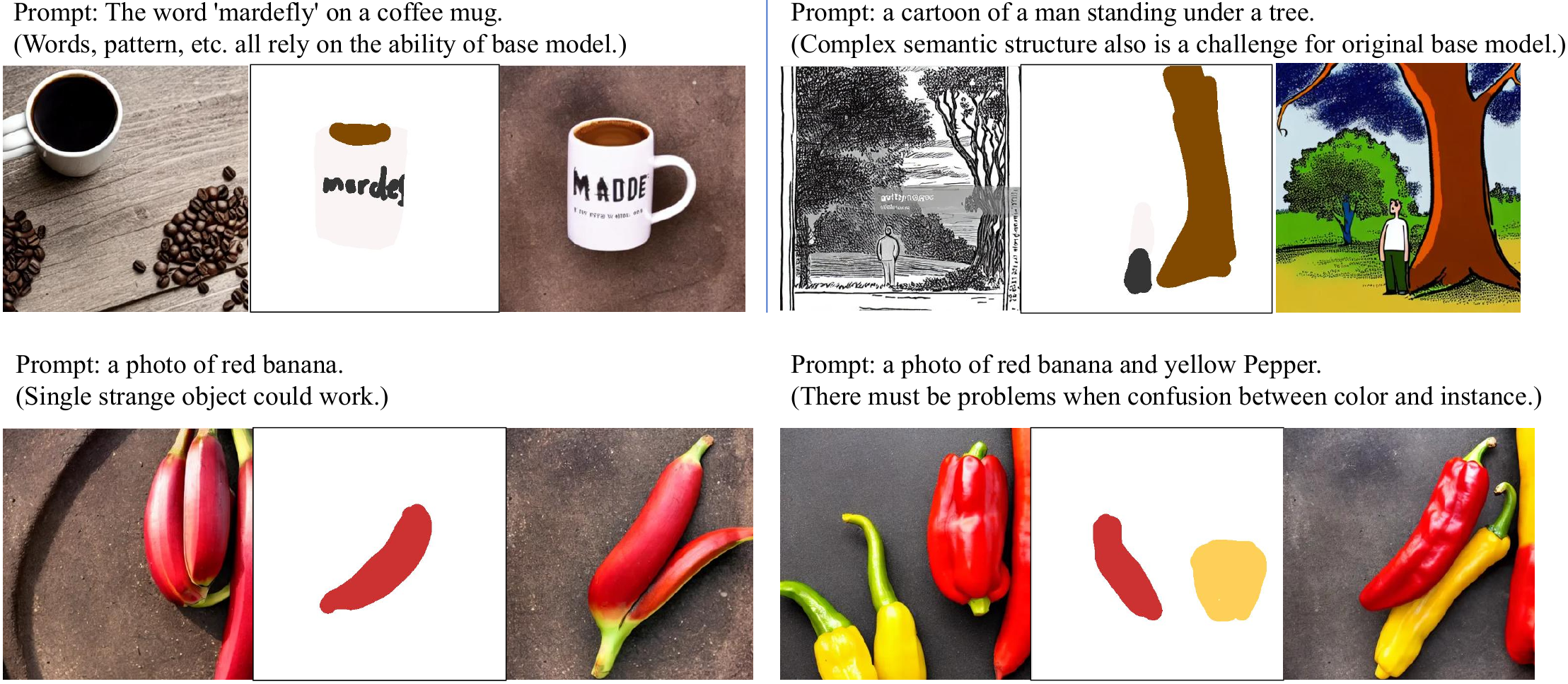}
    \caption{The visualization of failure cases, including complex texture, semantics, structure and confusion between color and object.}
    \label{fig:Failure}
\end{figure*}

\section{More ablative results}

\subsection{Quantitative Analysis of Individual guidance}

We provide a ablative study for individual guidance. As we know the effect of method or modules in image generation task is diffcult to quantitative. But we still conduct the quantitative analysis. We randomly sample 20 prompts from ImageNet-R-TI2I and draw 20 reference painting. In addition, we also add 10 prompts containing strange objects with confused color, such as "yellow apple and green banana". The results as follow:

\begin{table}[h]
\centering
\caption{\small{Quantitative Analysis of Individual guidance.}}
\small
\renewcommand\arraystretch{1.2}
\setlength{\tabcolsep}{2mm}
\begin{tabular}{lccccccc}
\toprule
 Color & Ins\&Sem & Latent Regeneration & CLIP score & LPIPS\\ \midrule 
 - & -   & -   & 0.268  & 0.239   \\
$\surd$ & -   & -   & 0.287  & 0.674   \\
- & $\surd$   & -   & 0.273  & 0.471   \\
$\surd$ & $\surd$   &  -  & 0.299  & 0.690   \\
- & -   & $\surd$   & 0.271  & 0.296  \\
$\surd$ & -   & $\surd$   & 0.293  & 0.682   \\
- & $\surd$   & $\surd$   & 0.273  & 0.508   \\
$\surd$ & $\surd$   & $\surd$   & 0.311  & 0.704   \\
\bottomrule
\end{tabular}\label{tab:user}
\end{table}

We do not agree the score would be the real effect on the generation task, but we wish the analysis result could be a help for future research. Although the score of Ins\&Sem and LR are not good, they are necessary to solve the problem about distinguishing between similar color instances, and confusion in assigning semantic attributes to text.

\subsection{Hyperparameter Analysis}
\begin{figure*}[h]
    \centering
    \includegraphics[width=0.915\linewidth]{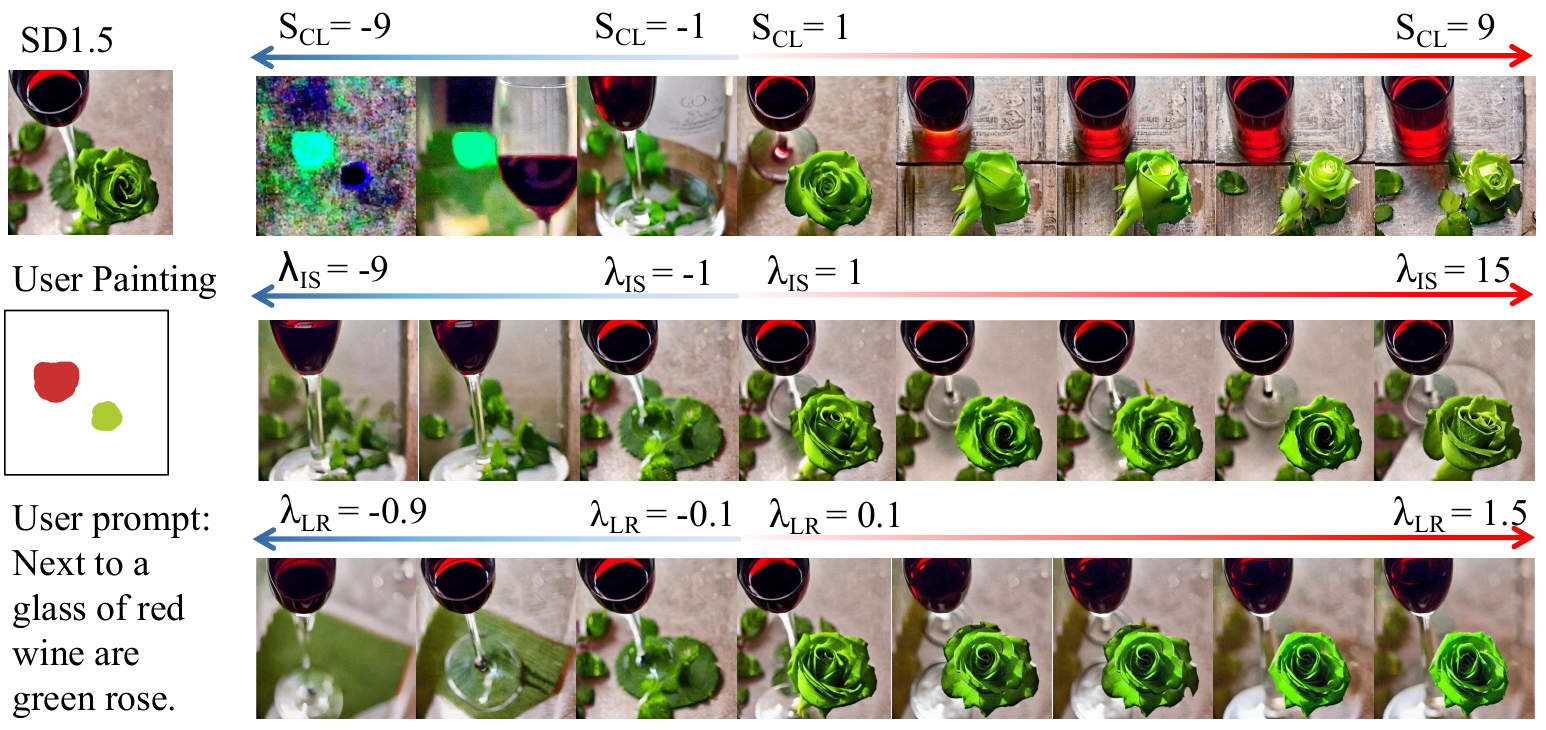}
    \caption{The influence of different guidance strength.}
    \label{fig:HA}
\end{figure*}
We also provide three independent hyperparameter ablation experiments with different guidance, as shown in Fig.~\ref{fig:HA}. We can see that when $S_{CL}$ is too large, although the instance maintains the correct color, its semantic representation begins to blur. When $S_{CL}$ is too small or even negative, the color structure of the generated image is disrupted. For $\lambda_{IS}$ and $\lambda_{LR}$, even if they become larger, they cannot guarantee color accuracy, but when they become smaller or negative, the corresponding semantic concepts may be removed from the graph.

\end{document}